\documentclass[11pt]{article}

\usepackage[margin=1in]{geometry}
\usepackage{booktabs}
\usepackage{amsmath}
\usepackage{graphicx}
\usepackage{url}
\usepackage[hidelinks]{hyperref}

\newcommand{\nm}{\ensuremath{\mathrm{N\,m}}}

\title{\textbf{Sim4RAD}: A MATLAB Toolkit for Design-Phase Sizing,\\ Actuator Selection and URDF-Model Verification\\ of Serial Robotic Arms}

\author{Jingchen Li\\ \small Tencent Robotics X Lab (affiliation when this work was conducted)\\ \small \texttt{jingchenli\_njzy@163.com}}

\date{2026}

\begin{document}

\maketitle

\begin{abstract}
Actuator sizing is an early and consequential decision: joint torque and velocity specifications determine which motors, gearboxes and brakes can be bought, yet they must be fixed before a CAD model, and therefore before the inertial parameters of the links, exist. The report presents \textbf{Sim4RAD} (Simulation for Robot Arm Design), a MATLAB toolkit for that phase and for the verification that follows it. It synthesizes a rigid-body model of a serial arm from a handful of parameters and drives every analysis from configuration files, so that a robot, a payload, a kinematic layout or a study is switched without editing code. Four analyses supply the specification: a \emph{static} analysis combining lever-arm sampling with an exhaustive posture search, which gives a \emph{rated} torque at a chosen payload envelope and a reproducible \emph{peak} torque; a \emph{dynamics} pipeline that reshapes a trajectory under one of four constraints, including the actuator torque--speed envelope, which reverses the question and asks how fast the selected modules can drive the arm; a \emph{stochastic} analysis over many random motions, so that a specification rests on a class of motions rather than on one trajectory; and a \emph{workspace and dexterity} analysis that separates \emph{where} the arm can reach from \emph{how} it can reach there. Replacing the synthesized model by a URDF file turns the same analyses to the verification direction.
A 7-DOF anthropomorphic arm is studied in three layouts that differ in the order of the two shoulder axes and in whether the last wrist joint is coaxial with the payload. At half arm length the static demand is nearly the same in all three, and it is this demand rather than the sampled motion that sizes the joints; reordering the shoulder joints leaves the worst-case torque unchanged at 55~\nm{} and alters the reachable volume by less than 8\,\%. A last wrist joint coaxial with the payload removes its gravity torque, reduces the adjacent wrist joint by 29\,\% and cuts the demand of the last joint by a factor of ten in a $5$~m/s scan, at the price of dexterity (the mean falls from 0.570 to 0.348). The same arm needs 20--37 times more torque for a $5$~m/s target than for a $2$~s motion: the scenario, not the trajectory, must be fixed first. The synthesized model is intended for sizing and comparison; verification should use the URDF model of the detailed design. The toolkit is available at \url{https://github.com/LasterLi/Sim4RAD}.
\end{abstract}

\section{Introduction}

Designing a robotic arm typically starts from a small set of requirements: degrees of freedom, reach, payload, and desired dynamic performance. The next step is to translate those requirements into joint specifications, because the motors, gearboxes and brakes must be selected long before a detailed mechanical design exists. At this point, however, neither the inertial parameters nor even the link geometry is known with any accuracy.

In practice, engineers either rely on hand calculations for a few assumed postures, or wait until a CAD model exists and then run a dynamics simulation. The first approach underestimates the worst case because it cannot explore the posture space; the second approach delays a decision that constrains everything downstream. Established dynamics libraries do not close this gap: they are excellent at computing the dynamics \emph{of} a given model, but they assume the model already exists.

This report describes \textbf{Sim4RAD} -- short for \emph{Simulation for Robot Arm Design} -- a MATLAB toolkit that starts from the requirement side, produces actuator specifications for the design phase, and later verifies the same arm on the URDF model exported from the detailed design. It is written as a technical report rather than as a research contribution: the individual numerical methods are standard, and the value of the work lies in the workflow, in the reproducibility of the case study, and in the availability of the implementation.

\subsection{Scope: two steps of a four-step flow}
\label{sec:scope}

It is useful to locate the toolkit in the overall design flow of an arm, which consists of four steps: requirements, scheme simulation and optimization, detailed design, and verification. Sim4RAD covers two of them: the second step, where no CAD model exists yet, and the fourth, where a detailed model does. Both are implemented rather than merely described -- the same library and the same entry scripts serve both, and only the model source is switched -- and the distinction between the two is not only the fidelity of the model but also the direction of the analysis:

\begin{itemize}
  \item \emph{Sizing} (step 2): given requirements, find the joint torque, velocity and workspace a candidate layout demands, and turn them into actuator specifications.
  \item \emph{Verification} (step 4): given the selected actuators and the detailed model, find what payload, what motion and what workspace the arm actually achieves, and check the requirements against it.
\end{itemize}

\begin{figure}[t]
\centering
\includegraphics[width=\linewidth]{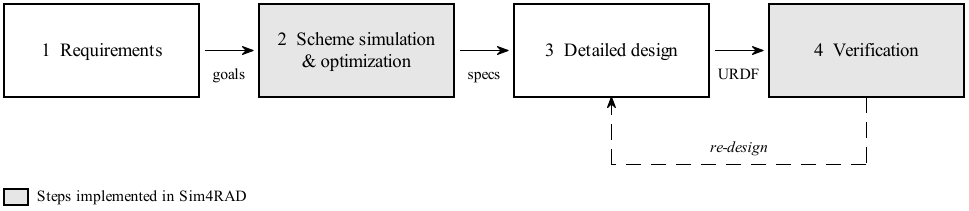}
\caption{The four steps of the arm design flow and the role of the toolkit. The shaded steps are implemented in Sim4RAD: scheme simulation and optimization, which works on a synthesized model because no CAD model exists yet, and verification, which works on the URDF model exported from the detailed design. The dashed arrow indicates that a failed requirement sends the design back to the detailed-design step.}
\label{fig:flow}
\end{figure}

Figure~\ref{fig:flow} illustrates this. The detailed-design step in between, in which the structure, transmissions and actuator selection are worked out in CAD software, is outside the scope of the toolkit.

\subsection{Contributions}

The specific elements of the toolkit are:

\begin{itemize}
  \item a parameterization that builds a rigid-body model of a serial arm \emph{before} CAD exists \cite{featherstone2008}, synthesizing link masses and inertias from an estimated total mass and per-link radii, with the same model later replaceable by a URDF model;
  \item a static analysis that (i) samples the joint space to relate the lever arm about the base to each joint torque, (ii) searches exhaustively over postures built from multiples of $90^{\circ}$ and the joint limits to obtain the worst case per joint, and (iii) merges postures with equal torque vectors to report a minimal set of ``typical'' postures;
  \item a dynamics pipeline that accepts a trajectory either as sampled data or as start/end states with joint-space or Cartesian interpolation, rescaled to satisfy one of four constraints: a target duration, a maximum end-effector velocity, a maximum end-effector acceleration, or the actuator envelope;
  \item a stochastic motion scan that runs the same pipeline over a large number of random motions and reports the torque and velocity demand as a percentile, so that a specification does not depend on one hand-picked trajectory;
  \item a workspace and dexterity evaluation that distinguishes reachability from the ability to approach a point from many directions;
  \item an optional transmission model that distinguishes joint-side and motor-side quantities through the gear ratio and a constant efficiency;
  \item an open-source implementation with a worked case study comparing three kinematic configurations of the same 7-DOF arm.
\end{itemize}

\section{Related work}
\label{sec:related}

\subsection{Rigid-body dynamics and simulation software}

Rigid-body dynamics algorithms are mature and well documented \cite{featherstone2008,siciliano2009}. Open implementations include the Robotics Toolbox for MATLAB \cite{corke2017}, Pinocchio \cite{carpentier2019}, RBDL \cite{felis2017} and general physics engines such as MuJoCo \cite{todorov2012}, PyBullet \cite{coumans2021} and Drake \cite{tedrake2019}. More recent platforms extend this base in two directions that matter for design: GPU-parallel simulation for large-scale sampling, such as Isaac Gym \cite{makoviychuk2021} and its successor Orbit \cite{mittal2023}, and differentiable simulators that expose gradients with respect to the dynamics \cite{werling2021}. These provide forward and inverse dynamics, Jacobians and, in several cases, Lie-group or gradient formulations suitable for optimization; they require the kinematic tree and the inertial parameters as input, and most of them are designed for control, estimation, learning or co-design rather than for the sizing problem addressed here.

\subsection{Kinematic design metrics and workspace analysis}

The metrics used to judge a kinematic design are equally classical. Manipulability \cite{yoshikawa1985} and Jacobian-based dexterity indices \cite{salisbury1982,klein1987,gosselin1991} quantify how well a posture can move and exert forces in all directions; they are used both as design objectives and as measures of distance from singularities. Task-oriented variants weight these measures by a specified task \cite{chiu1988}, and global indices integrate them over the workspace \cite{gosselin1991}. Workspace itself has been characterized geometrically since the early 1980s \cite{kumar1981}, and sampling-based estimates are now the practical route for redundant arms, where geometric boundaries are no longer sufficient. Design studies keep combining the two: a recent methodology optimizes a redundant dual-arm configuration against task-specific reachability and dexterity metrics \cite{guri2025}.

A recurring practical distinction, and one that this report relies on, is between \emph{reachability} (how far the arm can get) and \emph{dexterity} (how many ways it can get there). For a redundant 7-DOF arm both are needed, because redundancy buys posture freedom at a given point rather than reach.

\subsection{Load capacity and actuator sizing}

The quantities that this report derives are joint torques at a specified payload, which places it next to the literature on manipulator load capacity and on actuator selection. The dynamic load-carrying capacity problem -- the largest payload that a manipulator can carry along a trajectory without exceeding its actuator limits -- was formulated for serial arms by Wang and Ravani \cite{wang1988}. Posed at the design stage, it decides which components are feasible: a bi-level kineto-static formulation selects discrete components while checking reachability \cite{mingo2025}. On the design side, arm design has long been treated through dynamic quantities such as the generalized inertia ellipsoid \cite{asada1983} and the inertial properties of a manipulator in operational space \cite{khatib1995}, with a classical account of direct-drive arm design in \cite{asada1987}.

Actuator selection criteria themselves are treated in the mechatronics literature, for example through the inertia-matching and RMS-torque criteria used to pick a motor--gearbox pair for a duty cycle \cite{vandestraete1998}, and through the estimation models used for the preliminary design of electromechanical actuators in aerospace \cite{budinger2012}. The term \emph{actuator sizing} itself goes back to \cite{thomas1985}, which distributes the load capacity over the joints by local dynamic criteria; automated sizing frameworks are still being built, most recently for modular architectures with closed kinematic loops \cite{robodimm2026}. Task-based kinematic synthesis, in which link parameters are optimized against a set of specified tasks, was formulated by Paredis and Khosla \cite{paredis1993}. Standard documents also define the vocabulary: rated load and the associated test conditions for industrial manipulators are specified in ISO~9283 \cite{iso9283}.

The same trade-offs are treated in depth for dynamic legged machines, where torque density, impact mitigation, transparency and efficiency compete in the choice of actuator \cite{seok2015,wensing2017}, and a more recent line of work optimizes the morphology and the actuator parameters jointly by differentiating through a motion solver \cite{dinev2022}, a route that now extends to pretrained morphology-agnostic controllers \cite{strgar2025} and to the co-design of complete end-effectors \cite{fay2026}.

The toolkit implements a deliberate subset of this landscape. It does not optimize link lengths against a task, and it does not search for an actuator; it computes the demand quantities (rated and peak torque, peak velocity, posture set, workspace and dexterity) that these criteria and methods consume, on a model that exists before CAD. The converse problem -- optimizing a configuration and its drive train against such demands -- has been addressed in earlier work by the present author, for a service arm, by treating the layout, the workspace, the structural stiffness and the mass in one optimization \cite{li2021a,li2021b}, and the lightweighting that follows is a structural optimization problem of its own \cite{ding2021,li2026}.

\subsection{Anthropomorphic 7-DOF arms and design methodology}

Finally, the case study is an anthropomorphic 7-DOF arm, a configuration that appears repeatedly in research and collaborative robotics. Examples include the DLR lightweight robot and its descendants \cite{albuschaeffer2007}, a high-performance anthropomorphic arm designed for household applications \cite{liu2024}, and modular arms used in service robotics \cite{campeau2019}. The surveys of human--robot collaboration document why this class of arm exists and which requirements it inherits \cite{ajoudani2018,villani2018}, including the collision detection and reaction that shape its mechanical design \cite{haddadin2017}. The redundancy of such arms is usually discussed in terms of redundancy resolution and control \cite{siciliano1990}, while the kinematic layout choice -- shoulder ordering, wrist arrangement, offsets -- is normally fixed by mechanical and manufacturing considerations. The case study below isolates that choice quantitatively.

On the process side, the four-step flow of Section~\ref{sec:scope} mirrors systematic design methodology, which prescribes an explicit sequence from requirements to verification with iteration between the steps \cite{pahl2007}. The contribution of the toolkit is not the process but the ability to execute the second step with quantitative models that do not yet exist in CAD, and then to re-run the same analyses, unchanged, on the model that the detailed design produces (the fourth step).

\subsection{The gap addressed here}

The gap addressed here is therefore not algorithmic but procedural. A designer who has only a requirement sheet must first invent a model, and the quantities that matter at that stage -- the worst-case gravity torque per joint, the lever arm at which a payload should be specified, whether a joint is dominated by the payload or by the arm's own mass, and how much of the demand comes from inertia rather than gravity -- are properties of a posture space and of a set of candidate motions, not of a trajectory. General-purpose libraries answer a different question, actuator-selection criteria and load-capacity analyses assume that the demand is already known, and kinematic design metrics are usually evaluated on models that already exist. Sim4RAD fills the step in between, treating general-purpose dynamics algorithms as a dependency rather than a contribution.

\section{Toolkit overview}

The toolkit is organized around YAML configuration files; no script needs to be edited to change a robot or an analysis. Table~\ref{tab:params} summarizes the parameter sets.

\begin{table}[t]
\centering
\caption{Configuration files and their role.}
\label{tab:params}
\begin{tabular}{@{}ll@{}}
\toprule
File & Content \\
\midrule
\texttt{config\_modeling.yaml} & gravity direction, model source (manual or URDF), \\
                              & joint limits, end load, sampling count \\
\texttt{config\_manual\_robot.yaml} & DOF, link/joint frames, link radii, estimated mass \\
\texttt{config\_dynamics.yaml} & time step, planning space, interpolation, trajectory \\
                              & definition, constraint mode, actuator parameters \\
\texttt{config\_stochastic\_dynamics.yaml} & number of stochastic trajectories, parallelism \\
\texttt{config\_workspace.yaml} & space bounds, resolutions, base pose \\
\bottomrule
\end{tabular}
\end{table}

The functions are grouped by topic: \texttt{modeling}, \texttt{kinematics}, \texttt{trajectory}, \texttt{dynamics}, \texttt{static}, \texttt{workspace}, \texttt{interpolation}, \texttt{transform}, \texttt{visualization}, \texttt{math}, \texttt{arrangement}, \texttt{progress} and \texttt{config}. Each analysis is driven by an entry script (\texttt{script\_*.m}), so that the flow from configuration to figure is explicit and inspectable. The analyses share one robot object, so that the layout, the payload and the actuator parameters used by the static, dynamic and workspace studies are guaranteed to be the same.

\section{Methods}

\subsection{Modeling without CAD}
\label{sec:modeling}

A chain of $n$ revolute joints is defined by $n$ frames, each given relative to its parent by six parameters $[x, y, z, \phi, \theta, \psi]$ (position and ZYX Euler angles, in meters and degrees), with rotation about the local $z$ axis. Only revolute and fixed joints are supported; fixed joints allow an arbitrary number of links for $n$ degrees of freedom.

Inertial parameters are synthesized from two inputs: the estimated total mass $m_{\mathrm{tot}}$ and a radius $r_i$ per link. At the home configuration, the distance $L_i$ between consecutive frame origins gives an approximate cylinder of volume $V_i = \pi r_i^2 L_i$; the density $\rho = m_{\mathrm{tot}} / \sum_i V_i$ then assigns mass $m_i = \rho V_i$ to each link, with the center of mass at the midpoint of the segment and a cylinder inertia tensor
\begin{equation}
I_0 = \mathrm{diag}\left(\tfrac{1}{12} m_i (3 r_i^2 + L_i^2),\ \tfrac{1}{12} m_i (3 r_i^2 + L_i^2),\ \tfrac{1}{2} m_i r_i^2\right),
\label{eq:cylinder}
\end{equation}
rotated into the link frame and shifted to the center of mass by the parallel axis theorem. The same cylinder is used as collision geometry. The end load is appended as a final fixed body with a specified mass, offset and scalar inertia.

This construction is deliberately coarse. It is accurate enough to rank configurations and to bound actuator requirements, but it is not a substitute for the inertial data of a detailed design. The same pipeline accepts a URDF model once one exists, which is the mechanism by which the same analysis is reused for verification; the inertial data then comes from the model file, and the synthesized quantities of Eq.~\eqref{eq:cylinder} are not used at all.

\subsection{Static analysis}
\label{sec:static}

The static torque of each joint is computed by gravity torque with the arm at rest. Two complementary sweeps are performed.

\paragraph{Random sampling.} A number of configurations (20000--50000 in practice) are drawn uniformly from the joint box, and for each one the gravity torque of every joint is recorded together with a lever-arm metric: the horizontal distance $\| (x, y) \|$ of the end-point frame from the base axis. Plotting torque against this metric separates the joints whose demand grows with extension from those that do not, and the maximum over the subset below a chosen lever arm gives the payload requirement at that lever arm. This is the basis for distinguishing ``half-arm-length'' from ``full-arm-length'' payload specifications, and it is the quantity used for the continuous-duty (rated) torque.

\paragraph{Exhaustive search over structured postures.} Because the random sweep does not reach the corners of the joint box reliably, a second sweep enumerates, for each joint, the candidate angles $\{$lower limit, multiples of $90^{\circ}$ inside the limits, upper limit$\}$, and evaluates the Cartesian product of these sets. For the case study below this yields 14400 postures for layouts A and B and 24000 for layout~C, whose last joint has a symmetric $360^{\circ}$ range and therefore admits one more candidate angle. The resulting torques are exact for the enumerated set; in particular, postures where a link is horizontal are included, which is where the gravity lever arm is maximal for a serial arm. This sweep gives the momentary (peak) torque and is reproducible to the last digit.

\paragraph{Merging to typical postures.} Many enumerated postures are statically equivalent (for example, joint angles that move a mass along its own gravity axis). Postures whose torque vectors agree within $10^{-3}$ are merged, keeping the one closest to the home configuration. For each joint, the posture maximizing that joint's torque is retained, preferring the candidate with the largest overall torque norm, and a greedy cover selects a small subset of postures that jointly exhibit the worst case of every joint. The designer therefore receives a handful of postures to examine rather than thousands of numbers; these are the postures that must be checked for structural and packaging feasibility.

\subsection{Trajectories and dynamics}
\label{sec:dynamics}

A trajectory is defined in one of two ways. In the first, a table of samples is supplied (a CSV file); the planning space is either joint space (positions, optionally velocities, per row) or Cartesian space (position and quaternion, optionally linear and angular velocity), in which case inverse kinematics is solved per row. The time step is taken from the data.

In the second, only the start and end states are given, in joint or Cartesian space, and the motion is generated by cubic or quintic interpolation. The orientation of a Cartesian motion is interpolated in Euler angles (the ZYX sequence used by the quaternion conversions), with the Euler rates obtained from the angular velocity through the corresponding Jacobian; the parameterization is singular at a pitch of $\pm 90^{\circ}$, where the toolkit falls back on a pseudo-inverse instead of failing. The quintic form assumes zero end accelerations, since only positions and velocities are specified. The nominal duration of this interpolated motion is arbitrary because the trajectory is subsequently rescaled: a scalar time factor is applied to obtain the duration that satisfies the active constraint, and the position, velocity and acceleration profiles are re-sampled on the original grid.

Constraints are selected by a single configuration entry and only the active one is used. \emph{Mode 1} scales the trajectory to a target duration. \emph{Mode 2} and \emph{mode 3} solve for the duration at which the maximum end-effector linear velocity or acceleration equals a specified bound, using the forward kinematics of the rescaled trajectory. \emph{Mode 4} solves for the duration at which every joint stays inside the actuator envelope. In that last case the typically hyperbolic torque--speed characteristic of a motor with a gearbox is approximated by a linear envelope through the peak torque and peak speed, and the constraint is evaluated by rotating the sampled $(\omega, \tau)$ points into the envelope frame and comparing the maximum normal coordinate against the envelope, which is monotone in the duration and therefore solvable by a root finder. Mode 4 is the natural formulation for verification, where the actuators are given and the question is how fast the arm can move; modes 1--3 are the natural formulation for sizing, where a candidate duty cycle is given and the torque is the unknown.

Joint torques follow from inverse dynamics with gravity and an optional external wrench,
\begin{equation}
\tau_{\mathrm{out}} = \mathrm{ID}(q, \dot q, \ddot q, f_{\mathrm{ext}}),
\end{equation}
computed with the robot model, and the motor-side quantities are recovered through the transmission model
\begin{equation}
\tau_{\mathrm{motor}} = J_m \rho\, \ddot q + \frac{\tau_{\mathrm{out}}}{\rho\, \eta}, \qquad
\omega_{\mathrm{motor}} = \rho\, \dot q,
\label{eq:transmission}
\end{equation}
with gear ratio $\rho$, rotor inertia $J_m$ and constant efficiency $\eta$; power is evaluated on both sides. Equation~\eqref{eq:transmission} is the motor-side form, which differs by a factor $\rho$ from the joint-side form often quoted in the literature; note also that no separate high-speed gear inertia is parameterized, so any such contribution must be folded into $J_m$. Both quantities are reported because they are needed for different decisions: the joint-side torque sizes the structure, the motor-side torque and speed size the motor and gearbox.

\subsection{Stochastic motion requirements}
\label{sec:stochastic}

A single trajectory cannot fix a specification: it may be unrepresentative of the intended use, and small changes in the motion change the inertial demand substantially. The toolkit therefore offers a scan over randomly generated motions, which converts the question ``what torque does this motion need'' into ``what torque is needed by 97.7\,\% of the motions of this type''.

Each sample consists of a start and an end configuration drawn uniformly from the joint box, rejected if the end-effector displacement between them is smaller than a specified minimum (otherwise near-stationary motions dominate the sample). The pair is connected by joint-space interpolation and rescaled so that the end-effector maximum velocity equals a specified bound, i.e.\ the same mode-2 constraint as above but applied per sample; alternatively a fixed duration or a fixed acceleration bound can be used. Forward kinematics and inverse dynamics then give, per sample, the maximum end-effector speed and acceleration, the maximum torque and power of every joint, and the motion duration and distance, all of which are recorded. Plotting a requirement against the motion distance exposes how the demand scales with the size of the motion, and the specification is read off as a high percentile ($97.72\,\%$, i.e.\ mean plus two standard deviations) of the subset of motions that exceed a chosen distance.

Two consequences are worth stating explicitly. First, the result is a specification \emph{for a class of motions}, not for a trajectory: the input to the analysis is a velocity or acceleration bound rather than a path. Second, the percentage is not a safety factor but a statement about the population; the number of samples must therefore be reported, because a percentile estimated from $N$ samples is uncertain at the level of $1/\sqrt{N}$.

\subsection{Workspace and dexterity}
\label{sec:workspace}

Workspace analysis samples the reachable space on a grid of positions within a user-specified box, intersected with a maximum arm length, together with a discretization of approach directions. Two evaluation modes are available. In the inverse-kinematics mode, each position--orientation sample is tested for a feasible, collision-free solution under a five-dimensional constraint (position of the end-point frame plus the direction of its $z$ axis) on a 7-DOF arm, so that some samples are infeasible by construction; the fraction of the sampled directions that can be solved at a position is reported as its \emph{dexterity}. In the forward-kinematics mode, the joint space (tens of millions of configurations) is sampled and the results are mapped onto the same grid, which is the practical route for pure reachability. The resulting occupancy set can be visualized as a point cloud or as a dexterity map, where the colour is the fraction above.

The distinction between the two modes is a distinction between two design questions: whether a point can be reached at all, and whether it can be reached in the postures a task requires. Both are needed, and for a redundant arm they can disagree substantially, as the case study shows.

\subsection{From analysis to actuator specification}
\label{sec:spec}

The analyses above are combined into the specification the designer needs: a \emph{rated} torque, taken as the worst case within the half-arm-length payload envelope, which is a continuous-duty requirement; a \emph{peak} torque, taken from the exhaustive posture search, which is momentary; a peak velocity, taken from the velocity capability or from a trajectory constrained by mode 4; and a workspace and dexterity description, which determines whether the arm can perform the intended tasks at all. Because the same toolkit evaluates the $(\omega,\tau)$ working point of a candidate motion against the envelope, the consistency of the selected actuator with the intended motion can be checked without leaving the design-phase model, and a change of layout or payload is a change of configuration file rather than of code.

\section{Case study: a 7-DOF anthropomorphic arm}
\label{sec:case}

\subsection{Model and configurations}

The case study is a 7-DOF anthropomorphic arm with a spherical shoulder, a two-joint elbow, and a two-joint wrist, with a shoulder offset of 0.1~m, an upper arm of 0.3~m, a forearm of 0.3~m and an end-effector offset of 0.1~m, so that the shoulder-to-wrist length is 0.6~m and the base-to-load length is 0.7~m. The estimated arm mass is 10~kg (excluding the load) and the payload is 5~kg. Gravity is $-z$, and the link radii are 0.04, 0.04, 0.04, 0.03, 0.03, 0.02 and 0.02~m from base to tip.

\begin{figure}[t]
\centering
\includegraphics[width=0.82\linewidth]{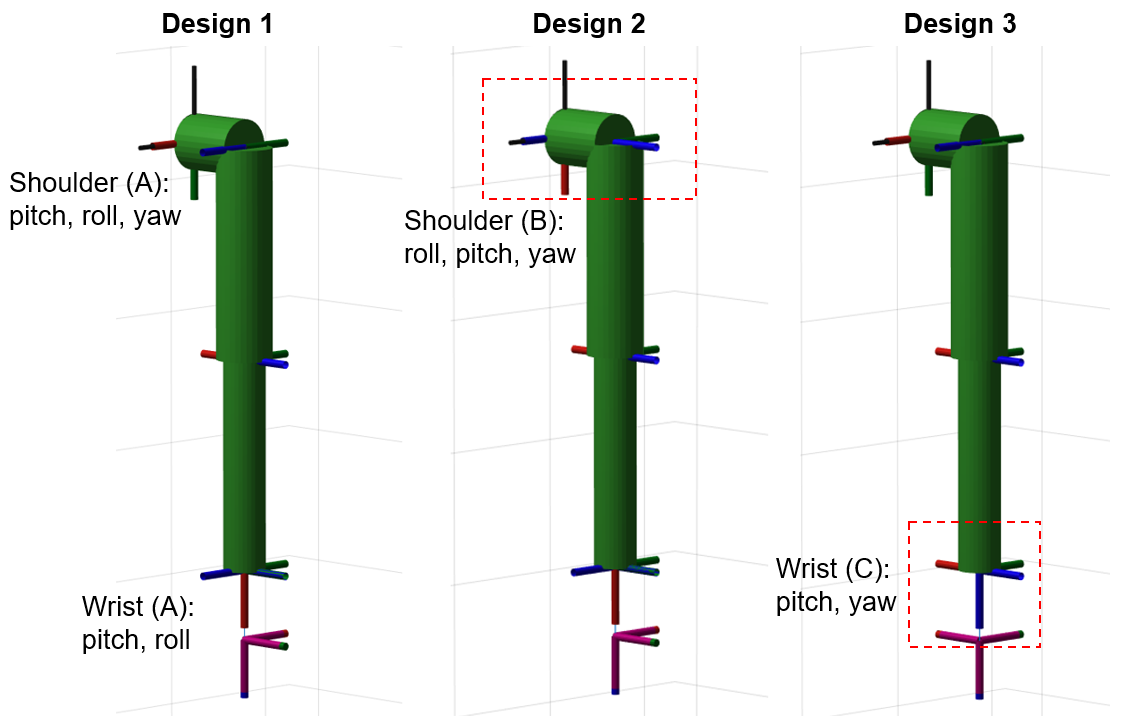}
\caption{The three layouts at the home configuration (arm hanging down, all joint angles zero); they are labelled Design 1--3 in the figure. All three have the same link lengths, the same shoulder offset and the same end load; they differ in the order of the first two shoulder axes (layout~B) and in the orientation of the last wrist axis relative to the payload (layout~C, where the last axis and the load offset are collinear).}
\label{fig:schemes}
\end{figure}

Three kinematic layouts are compared, shown in Figure~\ref{fig:schemes}. Layout~A orders the shoulder as pitch--roll--yaw; layout~B swaps the first two, giving roll--pitch--yaw; layout~C keeps layout~A but places the last wrist joint so that it is coaxial with the payload, with the load offset along the wrist axis. Table~\ref{tab:configs} lists the corresponding parameters.

\begin{table}[t]
\centering
\caption{The three kinematic layouts. Frames are relative to the parent frame as $[x,y,z,\phi,\theta,\psi]$ in m and degrees; limits in degrees. Layout~C shares the first six link frames of layout~A.}
\label{tab:configs}
\footnotesize
\setlength{\tabcolsep}{4pt}
\begin{tabular}{@{}lccc@{}}
\toprule
& A & B & C \\
\midrule
Link 1 & $[0,0,0,-90,0,0]$   & $[0,0,0,0,90,0]$    & as A \\
Link 2 & $[0,0,0.1,0,90,0]$  & $[0,0.1,0,-90,0,0]$ & as A \\
Link 3 & $[0,0,0,-90,0,0]$   & $[0,0,0,0,90,0]$    & as A \\
Link 4 & $[0,0,0.3,0,-90,0]$ & $[0,0,0.3,0,-90,0]$ & as A \\
Link 5 & $[0,0,0,0,90,0]$    & $[0,0,0,0,90,0]$    & as A \\
Link 6 & $[0,0,0.3,0,-90,0]$ & $[0,0,0.3,0,-90,0]$ & as A \\
Link 7 & $[0,0,0,90,0,0]$    & $[0,0,0,90,0,0]$    & $[0,0,0,0,90,0]$ \\
\midrule
Lower limits & $[-180,-45,-180,-150,$ & $[-45,-180,-180,-150,$ & $[-180,-45,-180,-150,$ \\
             & $\quad -180,-45,-90]$  & $\quad -180,-45,-90]$  & $\quad -180,-45,-180]$ \\
Upper limits & $[60,180,180,10,$      & $[180,60,180,10,$      & $[60,180,180,10,$ \\
             & $\quad 180,45,90]$     & $\quad 180,45,90]$     & $\quad 180,45,180]$ \\
\midrule
End load offset & $[0.1,0,0]$ & $[0.1,0,0]$ & $[0,0,0.1]$ \\
\bottomrule
\end{tabular}
\end{table}

Switching between layouts is done by activating configuration blocks, not by editing scripts; the layouts therefore share every other parameter, including the actuator capability. That capability is a set of seven commercial joint modules used as a reference throughout: peak torque $135$, $135$, $67.5$, $82.8$, $82.8$, $41.4$, $41.4$~Nm and peak speed $90$, $90$, $90$, $180$, $180$, $360$, $360$~rpm for joints $1\dots7$.

\subsection{Static torque requirements}

Table~\ref{tab:static} reports the static torque demand with the payload inside a 0.4~m lever arm (the half-arm-length requirement, sampled with 50000 configurations) and the worst case over the whole reachable set. Table~\ref{tab:maxstatic} reports the worst case found by the exhaustive posture search, which is reproducible to the last digit, unlike the sampled values.

\begin{table}[t]
\centering
\caption{Static torque demand with a 5~kg payload, in \nm. ``Half-arm'' is the maximum over sampled postures whose end-point horizontal distance is at most 0.4~m; ``full range'' is the maximum over all sampled postures. Sampling: 50000 configurations, uniform in the joint box.}
\label{tab:static}
\small
\begin{tabular}{@{}llrrrrrrr@{}}
\toprule
Layout & Requirement & J1 & J2 & J3 & J4 & J5 & J6 & J7 \\
\midrule
A & half-arm   & 36.66 & 38.14 & 23.44 & 23.86 & 4.89 & 4.88 & 4.90 \\
A & full range & 54.95 & 54.85 & 23.79 & 23.95 & 4.89 & 4.90 & 4.90 \\
B & half-arm   & 39.59 & 36.55 & 23.15 & 23.83 & 4.90 & 4.89 & 4.90 \\
B & full range & 57.14 & 55.02 & 23.79 & 23.90 & 4.90 & 4.90 & 4.90 \\
C & half-arm   & 36.46 & 38.48 & 23.62 & 23.95 & 3.44 & 4.90 & 0.00 \\
C & full range & 54.97 & 54.99 & 23.92 & 23.95 & 3.44 & 4.90 & 0.00 \\
\bottomrule
\end{tabular}
\end{table}

\begin{table}[t]
\centering
\caption{Worst-case static torque per joint from the exhaustive search over postures built from multiples of $90^{\circ}$ and the joint limits, in \nm. Layouts A and B give identical worst cases; only the wrist differs in layout~C.}
\label{tab:maxstatic}
\begin{tabular}{@{}lrrrrrrr@{}}
\toprule
Layout & J1 & J2 & J3 & J4 & J5 & J6 & J7 \\
\midrule
A & 55.14 & 55.14 & 23.96 & 23.96 & 4.90 & 4.90 & 4.90 \\
B & 55.14 & 55.14 & 23.96 & 23.96 & 4.90 & 4.90 & 4.90 \\
C & 55.14 & 55.14 & 23.96 & 23.96 & 3.46 & 4.90 & 0.00 \\
\bottomrule
\end{tabular}
\end{table}

\begin{figure}[t]
\centering
\includegraphics[width=\linewidth]{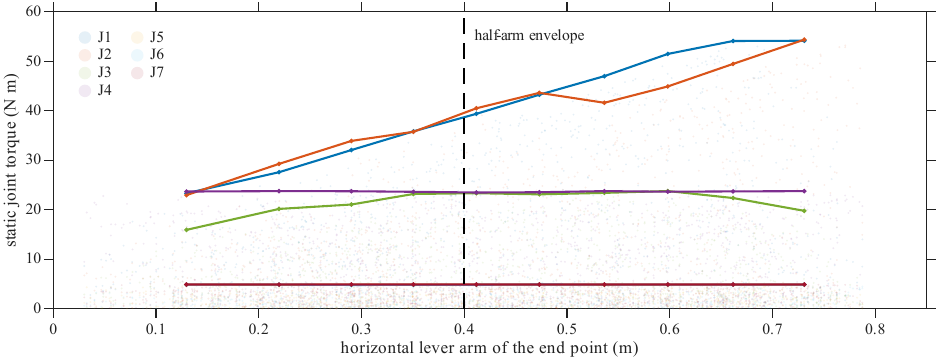}
\caption{Static torque against the horizontal lever arm of the end point, layout~A. Light clouds: individual random postures, one colour per joint (1000 of the 20000 samples are drawn, for legibility). Solid lines with markers: the maximum torque per joint within bins that hold an equal number of samples, i.e.\ the running requirement that defines the rated torque at a given payload envelope. The dashed line marks the half-arm-length envelope of 0.4~m used in Table~\ref{tab:static}. $J1$ and $J2$ grow with extension, $J3$ peaks and then declines, and the remaining joints are insensitive to extension. J5--J7 coincide at about 4.9~\nm.}
\label{fig:leverage}
\end{figure}

Figure~\ref{fig:leverage} shows the sampled sweep for layout~A: the clouds of individual postures, and the maximum torque within each lever-arm bin. Comparing the two rows of Table~\ref{tab:static} with each other, and the sampled worst case with the exhaustive one, gives three observations.

First, J1 and J2 grow nearly linearly with the lever arm, because both support the payload at a horizontal offset; the half-arm requirement is roughly two thirds of the full-range requirement for these joints. Their specification is therefore governed by the payload definition -- at which lever arm the payload must be supported -- and not by the arm's own mass.

Second, J3, the roll joint about the upper-arm axis, behaves differently: it rises, \emph{saturates}, and then declines as the arm approaches full extension (Figure~\ref{fig:leverage}). Its worst case is reached when the payload and the forearm mass have maximum lateral offset from the roll axis, which requires a bent elbow; straightening the arm removes that offset. Reading only the full-range maximum would suggest a demand nearly identical to the half-arm case, which is why the two rows are reported separately: for a joint of this type the choice of payload envelope changes the specification very little, and it is the other joints that decide where the envelope should be drawn.

Third, J4, J6 and J7 are essentially independent of extension: their worst cases are reachable at any lever arm, because their demand is set by the lateral offsets of the load and of the distal link masses rather than by the payload distance. Their magnitude is therefore insensitive to where the payload is specified, and for J5--J7 it is one tenth of the shoulder demand. This is the asymmetry that the layout change in the next section exploits.

\subsection{Effect of the layout choice}

The comparison in Table~\ref{tab:maxstatic} isolates what the layout choice does and does not change.

Reordering the shoulder (A $\to$ B) leaves the worst-case shoulder torques within about 4\,\% (57.14 against 54.95 for J1, 55.02 against 54.85 for J2); the exhaustively searched worst cases are in fact identical. The distribution of the demand between the two joints shifts (in layout~B the lateral swing joint sees a marginally higher worst case than the pitch joint, and the half-arm values change by up to 8\,\%), but no actuator changes size. For this arm, the shoulder ordering is therefore not a decisive design driver, which is itself a useful result: it means the choice can be made on other grounds, such as cable routing, the shoulder envelope or the reachable set at a given mounting.

The wrist layout (A $\to$ C) is decisive. Placing the last joint coaxial with the payload removes its gravity torque entirely (4.90 $\to$ 0.00), because the payload no longer has a lever arm about that axis, and it reduces the demand on the adjacent wrist joint by 29\,\% (4.90 $\to$ 3.46). Since the wrist actuators are at the end of the chain, where mass is most costly, this is the layout change with the clearest effect on the design. The same change has a cost, which the workspace analysis in Section~\ref{sec:wsresults} quantifies: the last joint can no longer change the direction of the tool, only its roll.

\subsection{One commanded motion}
\label{sec:motion}

The three layouts are then given the \emph{same} commanded motion, so that any difference in the result comes from the layout and not from the command. The motion is defined in joint space, from the home configuration to a target posture specified by functional joint angles -- shoulder pitch $45^{\circ}$, shoulder lateral $-30^{\circ}$, shoulder yaw $40^{\circ}$, elbow pitch $-60^{\circ}$, elbow roll $45^{\circ}$, wrist pitch $30^{\circ}$, last joint $45^{\circ}$ -- mapped onto the joint numbering of each layout, with zero start and end velocities and a cubic profile. Constraint mode 1 fixes the duration at 2~s. Because the commanded joint angles are the same, the position, velocity and acceleration profiles are identical for the three layouts; only the mapping onto physical axes differs. This is the control that makes the comparison meaningful, and the identity of the maximum joint velocities ($0.59$, $0.39$, $0.52$, $0.79$, $0.59$, $0.39$, $0.59$~rad/s for joints $1\dots7$ of layouts A and C, with the first two entries swapped in layout~B) verifies it.

\begin{figure}[t]
\centering
\includegraphics[width=\linewidth]{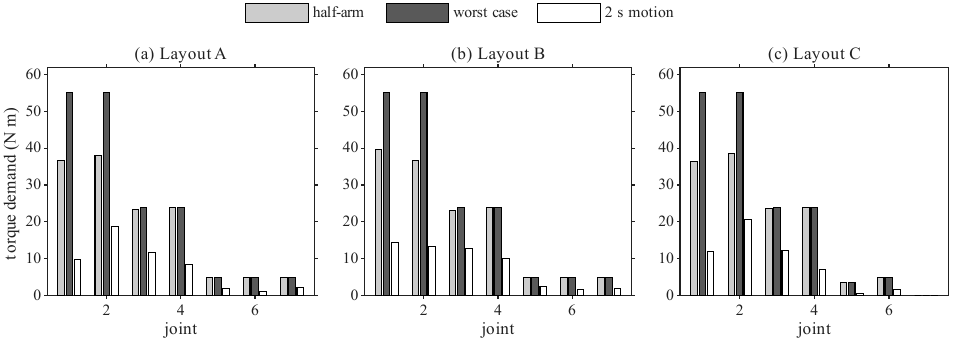}
\caption{Joint torque demand of the three layouts under three scenarios. Within each group of bars, from left to right: the half-arm-length requirement at a 5~kg payload, the worst case of the exhaustive posture search, and the maximum torque of the shared 2~s commanded motion of Section~\ref{sec:motion}. The joint index is the physical joint of each layout, so that layouts A and B differ in the assignment of J1 and J2. Values are recomputed by the plotting script.}
\label{fig:barrs}
\end{figure}

\begin{table}[t]
\centering
\caption{Maximum torque per functional joint over the shared 2~s commanded motion, in \nm. The last two columns report, for layout~A, the maximum gravity torque over all joints (identical trajectory with velocities and accelerations set to zero) and the maximum inertial torque; their maxima occur at different instants and therefore do not add up to the combined maximum of the A column.}
\label{tab:motion}
\small
\begin{tabular}{@{}lrrrrr@{}}
\toprule
Functional joint & A & B & C & A: gravity & A: inertia \\
\midrule
Shoulder pitch     & 9.67  & 13.34 & 11.99 & 20.66 & 2.46 \\
Shoulder lateral   & 18.82 & 14.41 & 20.54 &       &      \\
Shoulder yaw       & 11.77 & 12.87 & 12.10 &       &      \\
Elbow pitch        & 8.29  & 9.96  & 7.16  &       &      \\
Elbow roll         & 1.99  & 2.36  & 0.57  &       &      \\
Wrist pitch        & 1.10  & 1.58  & 1.55  &       &      \\
Last joint         & 2.08  & 1.99  & 0.02  &       &      \\
\midrule
Maximum & 18.82 & 14.41 & 20.54 & 20.66 & 2.46 \\
\bottomrule
\end{tabular}
\end{table}

Table~\ref{tab:motion} and Figure~\ref{fig:barrs} give the result. Four points are worth extracting.

\emph{The inertial term is small at this speed.} For layout~A the maximum torque along the trajectory is 18.8~\nm, while the gravity term alone reaches 20.7~\nm and the inertial contribution 2.5~\nm, i.e.\ 12--17\,\% of the gravity term across the layouts. The two do not simply add, because the maxima occur at different instants: the inertial term partly cancels gravity in the middle of the motion, so the gravity maximum is slightly higher than the combined maximum. The practical reading is that for a motion of this duration (2~s for joint excursions of $45^{\circ}$--$60^{\circ}$), \emph{the specification is set by the static analysis, not by the motion}. The maximum joint power is 7~W, and the largest per-joint utilisation of the reference modules is 19\,\% (J3 of layout~B); the global maximum uses 14\,\% of the shoulder capability.

\emph{The shoulder ordering redistributes rather than adds.} Between layouts A and B the shoulder pitch torque rises from 9.67 to 13.34~\nm while the lateral torque falls from 18.82 to 14.41~\nm; their sum changes by only 2.6\,\%. Since both joints are of the same size in this arm, the layout choice is immaterial for the actuators and can be made on mechanical grounds.

\emph{The coaxial wrist is visible even in a gravity-dominated motion.} The last joint of layout~C sees 0.02~\nm against 2.08~\nm in layout~A, and the elbow roll drops from 1.99 to 0.57~\nm. The price is paid at the shoulder, where the lateral torque rises from 18.82 to 20.54~\nm, because the payload offset is now axial rather than lateral.

\emph{The same joint command produces different end-effector motion.} With identical joint profiles, the maximum end-effector speed is 0.389, 0.440 and 0.414~m/s for A, B and C, and the displacement is 0.446, 0.502 and 0.492~m. Starting from $(0,0.100,-0.700)$~m, the end point finishes at $(-0.017,-0.261,-0.438)$, $(-0.114,-0.312,-0.437)$ and $(-0.070,-0.315,-0.446)$~m respectively. Layout~B moves the tool about 12\,\% further for the same joint command. This is a reminder that ``the same trajectory'' is ambiguous once the axis assignment changes; the joint-space definition removes that ambiguity for the actuators, but the task-space consequence remains.

\subsection{Stochastic motion requirements}
\label{sec:stochresults}

The single motion of the previous section cannot answer what the arm must be specified for, because the answer depends entirely on how fast the arm is expected to move. The requirements at the start of the design named a maximum end-effector speed of 5~m/s, so the stochastic scan of Section~\ref{sec:stochastic} is run at that bound: 100 random motions per layout, joint-space interpolation, each rescaled so that its end-effector speed peaks at 5~m/s, with a minimum end-effector displacement of 0.1~m for a motion to be sampled. The scans take 0.09--0.96~s (median 0.37~s) and cover a median displacement of 0.6~m, i.e.\ nearly the full extent of the arm.

\begin{table}[t]
\centering
\caption{Torque demand at the 97.72th percentile of random motions whose displacement exceeds 0.2~m, with each motion rescaled to a 5~m/s end-effector peak, in \nm. The last column is the peak torque of the reference joint modules.}
\label{tab:stochtorque}
\begin{tabular}{@{}lrrrr@{}}
\toprule
Joint & A & B & C & Limit \\
\midrule
1 & 404.61 & 426.72 & 403.73 & 135 \\
2 & 628.01 & 587.70 & 702.74 & 135 \\
3 & 315.25 & 205.62 & 294.14 & 67.5 \\
4 & 263.38 & 329.96 & 308.41 & 82.8 \\
5 & 56.89  & 59.18  & 46.27  & 82.8 \\
6 & 65.70  & 56.91  & 78.43  & 41.4 \\
7 & 58.27  & 70.55  & 6.25   & 41.4 \\
\bottomrule
\end{tabular}
\end{table}

\begin{table}[t]
\centering
\caption{Joint speed demand at the 97.72th percentile under the same 5~m/s scans, in rpm. The last column is the peak speed of the reference joint modules.}
\label{tab:stochvel}
\begin{tabular}{@{}lrrrr@{}}
\toprule
Joint & A & B & C & Limit \\
\midrule
1 & 200.95 & 162.62 & 203.38 & 90 \\
2 & 113.29 & 141.92 & 121.80 & 90 \\
3 & 210.75 & 192.72 & 213.22 & 90 \\
4 & 122.51 & 126.85 & 103.42 & 180 \\
5 & 219.89 & 344.22 & 264.70 & 180 \\
6 & 88.73  & 91.73  & 116.40 & 360 \\
7 & 135.97 & 181.39 & 372.52 & 360 \\
\bottomrule
\end{tabular}
\end{table}

\begin{figure}[t]
\centering
\includegraphics[width=0.85\linewidth]{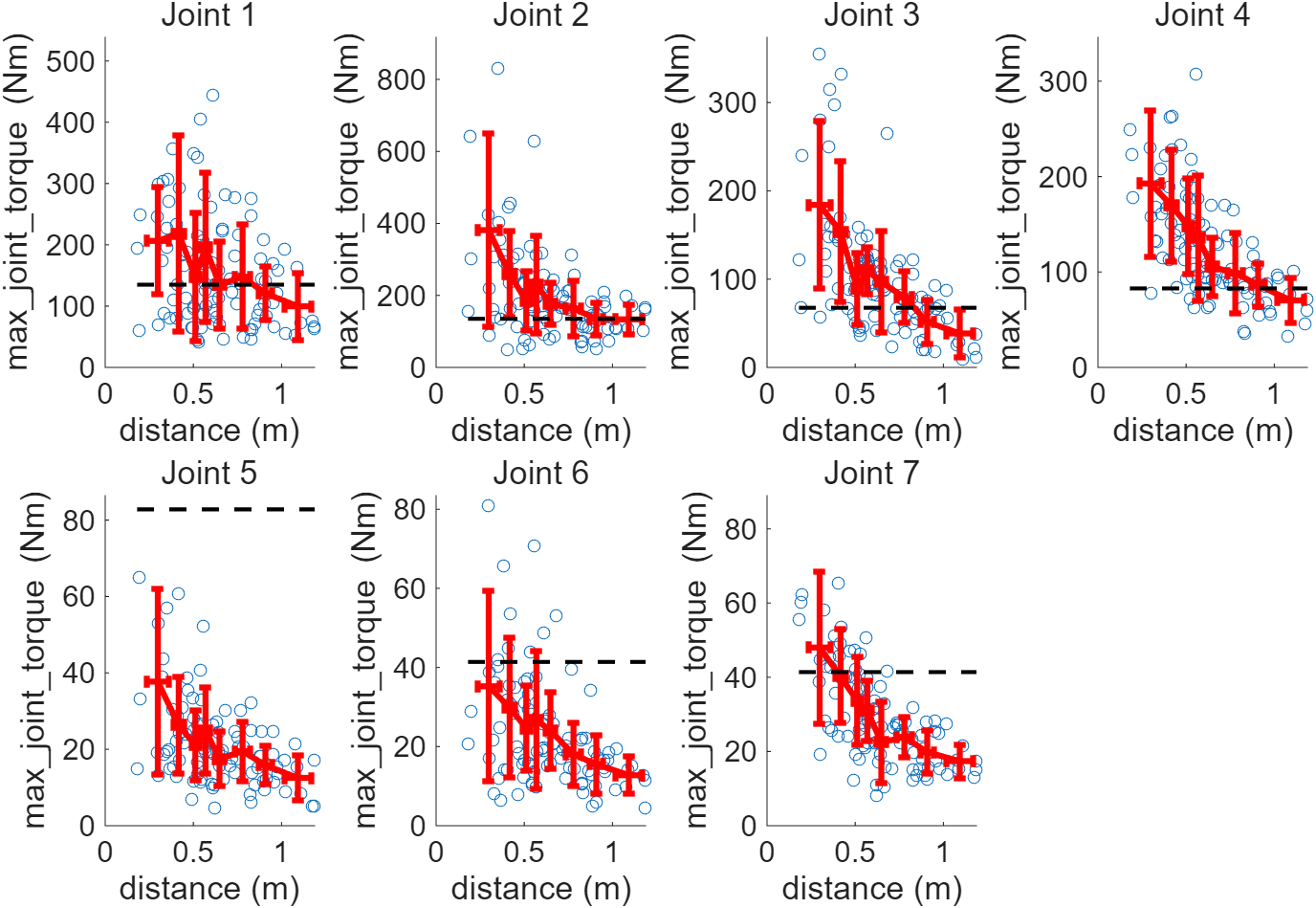}
\caption{Stochastic torque requirement of layout~A: one panel per joint, with the maximum torque of each random motion against the displacement of that motion, the mean and standard deviation per distance interval, and the peak capability of the corresponding reference module as a dashed line. All joints are above their capability at this end-effector speed.}
\label{fig:stoch}
\end{figure}

The outcome, in Tables~\ref{tab:stochtorque} and~\ref{tab:stochvel} and Figure~\ref{fig:stoch}, is a specification problem rather than a design trade-off. Shoulder torques of 404--703~\nm are three to more than five times the capability of the reference shoulder module, and the elbow joints require 206--330~\nm against 67.5--82.8~\nm. The wrist picture is mixed and worth stating precisely: J5 remains inside its 82.8~\nm capability at 46--59~\nm, J6 exceeds its 41.4~\nm in all three layouts (57--78~\nm), and J7 exceeds it in layouts A and B (58.3 and 70.6~\nm) but requires only 6.3~\nm in layout~C. Speeds are exceeded as well, at joints 1, 2, 3 and 5 and, in layout~C, at joint 7 (372.5~rpm against 360~rpm). Three conclusions follow.

First, at this speed the demand is entirely inertial: the same joints required 55~\nm in the worst static posture and 18.8~\nm in the 2~s motion, so the requirement has grown by a factor of about 7--13 relative to the worst static case and by 20--37 relative to the 2~s motion. Reaching 5~m/s within a 0.6~m displacement implies end-effector accelerations of 87--89~m/s$^2$, and joint accelerations in the hundreds of rad/s$^2$.

Second, the requirement of 5~m/s is inconsistent with the simultaneously stated acceleration limit of 10~m/s$^2$: accelerating to 5~m/s at 10~m/s$^2$ requires at least $v^2/(2a) = 1.25$~m of travel, while the arm's maximum extent is 0.7~m. The simulation makes the inconsistency quantitative: the motions it produces have peak accelerations about nine times the stated limit. This is precisely the situation that Section~\ref{sec:dynamics} predicts, namely that velocity and acceleration bounds cannot in general both be met and that only one of them can define the motion.

Third, layout~C is still better at the wrist, but the advantage must be stated correctly. The last joint requires 6.25~\nm against 58.3 and 70.6~\nm (a factor of ten), and the elbow roll drops by 19\,\%, yet the last joint is no longer negligible: at low speed its torque was zero because the payload has no lever arm about a coaxial axis, but at high speed the inertial coupling still loads it. \emph{``A coaxial joint carries no torque'' holds only in the gravity-dominated regime}, a caveat that no static analysis can reveal.

Finally, the differences between the layouts are mostly 10--20\,\%, of the same order as the sampling uncertainty at 100 samples, although J3 differs by up to 35\,\% (205.6 against 315.3~\nm). At this end-effector speed the layout is not what decides whether the design is feasible; the speed requirement is. If the question is instead how fast the reference modules can drive this arm, the formulation to use is mode 4 of Section~\ref{sec:dynamics}, which takes the actuator envelope as given and solves for the achievable duration, rather than assuming a speed and testing it.

\subsection{Workspace and dexterity}
\label{sec:wsresults}

The final analysis asks the complementary question: where can the tool go, and in how many orientations. The workspace is discretized at 0.15~m with 12 approach directions per position (336 grid positions inside the sampling box and the arm-length sphere), and each position--direction pair is tested by constrained inverse kinematics.

\begin{table}[t]
\centering
\caption{Workspace of the three layouts: reachable positions of the grid, reachable volume estimated by one voxel per position, and dexterity statistics. Dexterity is the fraction of the 12 sampled approach directions that admit a collision-free solution at a position.}
\label{tab:ws}
\begin{tabular}{@{}lrrr@{}}
\toprule
& A & B & C \\
\midrule
Reachable positions & 246 & 265 & 245 \\
Reachable volume (m$^3$) & 0.830 & 0.894 & 0.827 \\
Mean dexterity & 0.570 & 0.594 & \textbf{0.348} \\
Maximum dexterity & 1.000 & 1.000 & \textbf{0.667} \\
Positions with dexterity $\ge 0.5$ & 68\,\% & 73\,\% & \textbf{21\,\%} \\
\bottomrule
\end{tabular}
\end{table}

\begin{figure}[t]
\centering
\includegraphics[width=0.46\linewidth]{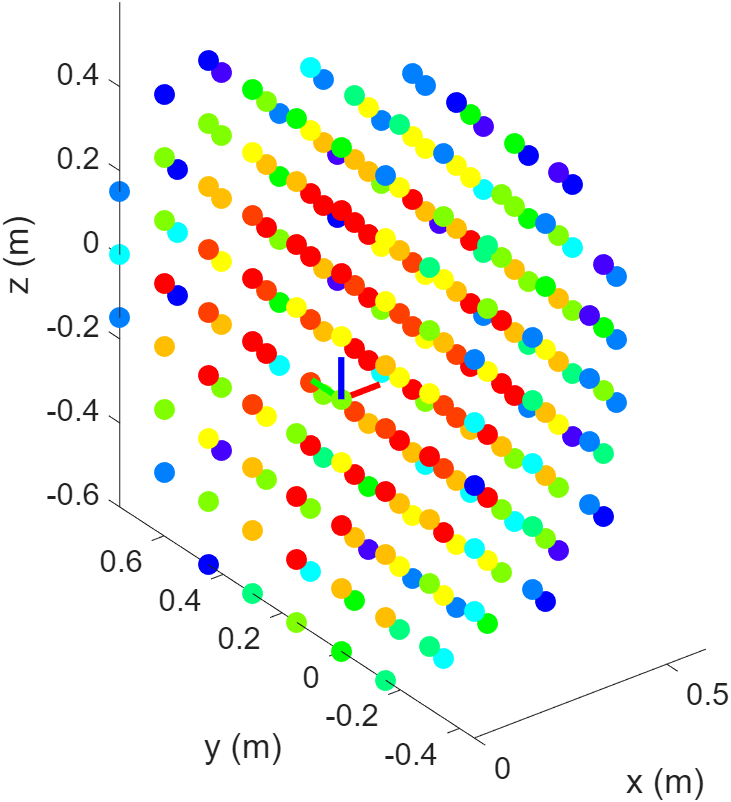}\hfill
\includegraphics[width=0.46\linewidth]{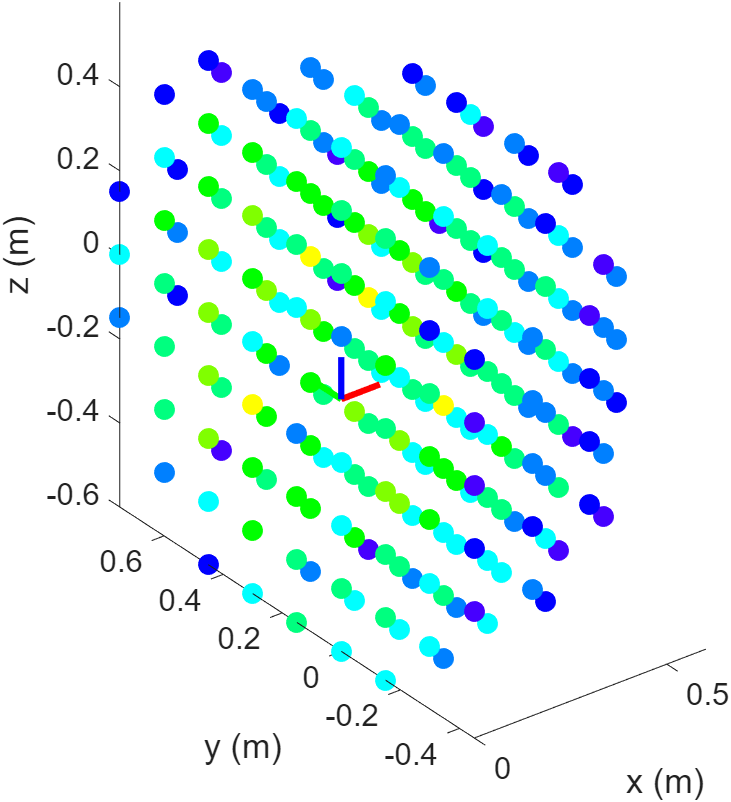}
\caption{Reachable workspace of layout~A (left) and layout~C (right), coloured by dexterity. The two layouts reach essentially the same volume, but layout~C admits far fewer approach directions at most positions.}
\label{fig:ws}
\end{figure}

Two results follow from Table~\ref{tab:ws} and Figure~\ref{fig:ws}. First, the layout barely affects \emph{where} the arm can reach: 246, 265 and 245 positions and volumes of 0.83, 0.89 and 0.83~m$^3$ differ by about 8\,\%, consistent with the static finding that reordering the shoulder changes little. Second, the layout strongly affects \emph{how} the arm can reach: the mean dexterity falls from 0.570 to 0.348 in layout~C, the maximum dexterity from 1.000 to 0.667, and the fraction of positions from which at least half of the sampled directions are feasible from 68\,\% to 21\,\%. The trade against Section~\ref{sec:motion} is now explicit: the coaxial wrist removes 29\,\% of the adjacent wrist torque and all of the gravity torque of the last joint, and in exchange it removes most of the wrist's ability to orient the tool, because the last joint can only roll about the forearm axis.

One caveat limits how strongly the absolute values of this comparison can be read, and it comes from the analysis rather than from the arm: the discretization is coarse, since dexterity can only take the values $k/12$ and the volume estimate treats each reachable position as a full voxel. The comparison itself, however, is like for like. Although the load offset is parameterized differently -- laterally with a $90^{\circ}$ rotation in layouts A and B, axially in layout~C -- all three layouts place the end-load frame at the same point and orientation at the home configuration, and the 12 approach directions are defined with respect to that frame in every layout. The dexterity loss of layout~C is therefore a genuine kinematic loss -- its last joint can only roll the tool about its own axis -- and not an artifact of the frame convention.

\subsection{Summary of the three layouts}

Table~\ref{tab:summary} collects the quantities that the case study produced. It is the table a designer would carry into the detailed-design step.

\begin{table}[t]
\centering
\caption{The three layouts side by side. Torque vectors are given for joints J1--J7 in \nm. ``Static half-arm'' and ``static worst case'' are the two static sweeps; ``2~s motion'' is the shared commanded motion of Section~\ref{sec:motion}; ``5~m/s scan, J7'' is the 97.72th percentile of the stochastic scan.}
\label{tab:summary}
\footnotesize
\setlength{\tabcolsep}{4pt}
\begin{tabular}{@{}lccc@{}}
\toprule
& A & B & C \\
\midrule
Difference from A & --- & shoulder order swapped & last joint coaxial \\
Static half-arm (\nm) & \shortstack[r]{36.7 / 38.1 / 23.4 / 23.9 /\\ 4.9 / 4.9 / 4.9} & \shortstack[r]{39.6 / 36.6 / 23.2 / 23.8 /\\ 4.9 / 4.9 / 4.9} & \shortstack[r]{36.5 / 38.5 / 23.6 / 24.0 /\\ 3.4 / 4.9 / 0} \\
Static worst case (\nm) & \shortstack[r]{55.1 / 55.1 / 24.0 / 24.0 /\\ 4.9 / 4.9 / 4.9} & same as A & \shortstack[r]{55.1 / 55.1 / 24.0 / 24.0 /\\ 3.5 / 4.9 / 0} \\
2~s motion, max (\nm) & 18.8 & 14.4 & 20.5 \\
5~m/s scan, J2 (\nm) & 628.0 & 587.7 & 702.7 \\
5~m/s scan, J7 (\nm) & 58.3 & 70.6 & 6.3 \\
Reachable volume (m$^3$) & 0.830 & 0.894 & 0.827 \\
Mean dexterity & 0.570 & 0.594 & 0.348 \\
\bottomrule
\end{tabular}
\end{table}

\section{Discussion}
\label{sec:discussion}

\subsection{What the case study establishes}

The results support four statements about this class of arm, from the most transferable to the most specific.

\emph{The scenario, not the trajectory, fixes the specification.} The same joints and payload require 55~\nm in the worst static posture, 18.8~\nm for a 2~s motion and 628~\nm for motions that reach 5~m/s, so a specification quoted without its scenario is meaningless, and the first decision in the sizing step is which of these regimes the arm is for.

\emph{Statics governs moderate-speed sizing, inertia governs high-speed sizing.} At a 2~s motion the inertial term is 12--17\,\% of the gravity term and statics is the binding constraint; the crossover lies where the acceleration times the effective inertia exceeds the gravity torque, of the order of ten rad/s$^2$ for this arm, well below the hundreds of rad/s$^2$ reached at 5~m/s.

\emph{The wrist is where a layout decision has leverage, and it is a trade rather than an improvement.} A last joint coaxial with the payload is the only change studied here that alters the size of an actuator: it removes that joint's gravity torque, cuts the adjacent wrist joint by 29\,\% and lowers the last joint demand tenfold at 5~m/s. The price is the wrist's ability to change tool direction (mean dexterity 0.570 $\to$ 0.348, and positions with at least half the directions feasible 68\,\% $\to$ 21\,\%). It is the right choice when the task fixes the tool direction and a light wrist matters, the wrong one when a point must be approached from many directions, and neither analysis alone shows both sides.

\emph{Shoulder ordering is not a sizing variable for this arm.} Swapping the first two shoulder axes leaves the worst-case torque identical, redistributes the half-arm requirement by less than 8\,\% and changes the reachable volume by less than 8\,\%. The choice can therefore be made on other grounds -- cable routing, the shoulder envelope, the mounting interface -- which removes a variable from the search.

\subsection{Which analysis governs which specification}

The case study also shows how the four analyses divide the labour; using the wrong one gives a plausible but wrong number.

\begin{itemize}
  \item the \emph{sampled static sweep} gives the rated torque -- the quantity to compare with a catalogue rating -- and its dependence on the lever arm is what makes the payload envelope a design decision;
  \item the \emph{exhaustive posture search} gives the peak torque and the typical postures that the structure and the packaging must accommodate, reproducibly, which matters when a specification is negotiated;
  \item the \emph{trajectory analysis} gives what a given duty cycle requires, and it is the only analysis that separates gravity from inertia; with mode 4 it also answers how fast the selected actuators allow the arm to move;
  \item the \emph{stochastic scan} gives the requirement of a class of motions and its sensitivity to the motion regime -- this is the analysis that would have caught the inconsistency between the 5~m/s and 10~m/s$^2$ requirements before an actuator was selected;
  \item the \emph{workspace and dexterity analysis} tells whether the tasks are possible at all and, for a redundant arm, what a wrist simplification costs.
\end{itemize}

\subsection{Relation to existing tools and methods}

The toolkit does not compete with general dynamics libraries, and the case study benefits from treating them as a dependency: the dynamics of a rigid-body tree is a solved problem, and the open implementations of Section~\ref{sec:related} are better tested than anything a toolkit of this kind should reimplement. What is added is the model construction that precedes them (a rigid-body model from estimated parameters, replaceable by URDF later), the sweep-and-merge logic that turns a posture space into a few specifications, and the configuration-driven coupling of statics, dynamics and workspace on one model.

Relative to the load-capacity and actuator-selection literature \cite{wang1988,vandestraete1998,budinger2012} the toolkit stops one step earlier: it produces the demand those criteria consume rather than selecting a part, so the link to a catalogue item is a comparison of numbers, not an optimization -- selection depends on catalogue data, duty cycles, thermal behaviour and cost, which a toolkit of this kind cannot have. Feeding the computed $(q, \dot q, \ddot q)$ histories into an RMS-torque and thermal criterion and searching over gear ratios, which the transmission model already parameterizes, is the natural extension. Relative to task-based kinematic synthesis \cite{paredis1993} and global dexterity optimization \cite{gosselin1991}, the toolkit evaluates candidate layouts instead of optimizing link lengths; the case study supports that division, since what dominated the outcome was discrete layout choices rather than continuous lengths, which a parametric optimizer would not have found.

\subsection{Threats to validity}

The results above should be read as sizing and comparison numbers, not as verification; verification requires the URDF model of a detailed design. Specifically:

\begin{itemize}
  \item link masses and inertias are synthesized from an estimated total mass and cylinder geometry, the end load is a point mass with a scalar inertia, and gravity is a constant vector; with the center of mass at the segment midpoint, individual link inertias of a real design may differ by tens of percent, so the ranking of layouts is more robust than the absolute torques;
  \item friction, damping, backlash and compliance are represented only by a constant transmission efficiency and a rotor inertia;
  \item static requirements from random sampling fluctuate with the number of samples: the values in Table~\ref{tab:static} come from a single run with 50000 configurations and \texttt{rng(1)} set beforehand, the entry script itself not fixing the seed. Table~\ref{tab:maxstatic} is reproducible but exact only for the enumerated posture set, since the grid contains multiples of $90^{\circ}$ rather than the true worst case;
  \item the stochastic requirements are percentiles of 100 motions per layout, so a 97.72\,\% percentile is estimated from the largest few samples: Tables~\ref{tab:stochtorque} and~\ref{tab:stochvel} should be read as ``about this magnitude'', and layout differences below about 20\,\% at that speed are not significant. The sample count was limited by runtime, not by the method;
  \item Cartesian-space planning solves inverse kinematics at every sample and is therefore sensitive to singularities and to the initial guess; and self-collision checking, which relies on the same coarse cylinder geometry, is unusable with the synthesized model of this arm -- neighbouring cylinders touch at zero-length links -- so it is meaningful only once a URDF model exists;
  \item the workspace comparison uses the same end-frame convention across layouts (Section~\ref{sec:wsresults}) but a coarse discretization, so its ordering is robust and its absolute dexterity values are not; and the case study covers a single arm family, a single payload and one commanded motion per analysis, with no experimental validation and no cross-check against another dynamics implementation.
\end{itemize}

\subsection{Outlook: closing the loop}

Figure~\ref{fig:flow} has a second half that this report does not exercise, although the toolkit implements it: with the model switched to a URDF file, and nothing else changed, the same entry scripts produce the verification-side quantities -- the payload a selected module set can carry, the motion it can drive under the actuator envelope (mode 4), and the workspace that is genuinely reachable once self-collision is meaningful. Three extensions would strengthen that path: cross-validating the dynamics against an independent implementation (Pinocchio or MuJoCo) on the same URDF, which would bound the current implementation risk; adding structural compliance and joint friction, the two largest omissions of the synthesized model; and adding a thermal and RMS-torque criterion, which would complete the chain from demand to a selected part. None of them changes the methods described here.

\section{Reproducibility}
\label{sec:reproducibility}

Every number in Tables~\ref{tab:static}--\ref{tab:summary} is produced by an entry script from a configuration file, and the layout is switched by activating the corresponding configuration blocks rather than by editing code:

\begin{itemize}
  \item the static tables by the random sampling entry script (50000 configurations; the reported values are from a run with \texttt{rng(1)} set beforehand, as the script itself does not fix the seed) and the exhaustive search entry scripts;
  \item the commanded motion of Section~\ref{sec:motion} by the dynamics entry script with constraint mode 1 and a duration of 2~s;
  \item the stochastic tables by the stochastic analysis entry script with constraint mode 2 and a 5~m/s bound, followed by the visualization script, which reads the stored result and computes the percentiles;
  \item the workspace table by the workspace analysis and visualization scripts at the resolution stated in Section~\ref{sec:wsresults};
  \item the verification-side quantities -- the payload a selected module set can carry, the duration it achieves under the actuator envelope (mode 4), and the workspace that is genuinely reachable once self-collision is meaningful -- by the same entry scripts with the model switched to a URDF file (\texttt{using\_urdf\_model} in \texttt{config\_modeling.yaml}).
\end{itemize}

The figures of this report are generated from the same sources: Figure~\ref{fig:flow} is drawn by a standalone script, the remaining figures come from the visualization entry scripts, and the two figures that are not produced by the toolkit's own plotting routines (\ref{fig:leverage} and~\ref{fig:barrs}) are drawn by scripts that recompute the underlying quantities rather than embedding them, so that a change of configuration or of model is reflected in those figures as well. The case study document shipped with the repository contains the same analyses with the full set of figures and the intermediate results; it is provided in English (\texttt{docs/case\_study.md}) and in Chinese (\texttt{docs/case\_study\_CN.md}).


\begin{thebibliography}{9}

\bibitem{ajoudani2018}
A.~Ajoudani, A.~M.~Zanchettin, S.~Ivaldi, A.~Albu-Sch\"affer, K.~Kosuge, and O.~Khatib,
``Progress and prospects of the human--robot collaboration,''
\emph{Autonomous Robots}, vol.~42, no.~5, pp.~957--975, 2018.

\bibitem{albuschaeffer2007}
A.~Albu-Sch\"affer, S.~Haddadin, C.~Ott, A.~Stemmer, T.~Wimb\"ock, and G.~Hirzinger,
``The DLR lightweight robot: design and control concepts for robots in human environments,''
\emph{Industrial Robot}, vol.~34, no.~5, pp.~376--385, 2007.

\bibitem{asada1983}
H.~Asada,
``A geometrical representation of manipulator dynamics and its application to arm design,''
\emph{ASME J. Dynamic Systems, Measurement, and Control}, vol.~105, no.~3, pp.~131--135, 1983.

\bibitem{asada1987}
H.~Asada and K.~Youcef-Toumi,
\emph{Direct-Drive Robots: Theory and Practice}.
MIT Press, 1987.

\bibitem{budinger2012}
M.~Budinger, J.~Liscou\"et, F.~Hospital, and J.-C.~Mar\'e,
``Estimation models for the preliminary design of electromechanical actuators,''
\emph{Proc. IMechE Part G: J. Aerospace Engineering}, vol.~226, no.~3, pp.~243--259, 2012.

\bibitem{campeau2019}
A.~Campeau-Lecours, H.~Lamontagne, S.~Latour, P.~Fauteux, V.~Maheu, F.~Boucher, C.~Duchaine, and C.~Gosselin,
``Kinova modular robot arms for service robotics applications,''
in \emph{Rapid Automation: Concepts, Methodologies, Tools, and Applications}, IGI Global, 2019.

\bibitem{carpentier2019}
J.~Carpentier, G.~Saurel, G.~Buondonno, J.~Mirabel, F.~Lamiraux, O.~Stasse, and N.~Mansard,
``The Pinocchio C++ library: a fast and flexible implementation of rigid body dynamics algorithms and their analytical derivatives,''
in \emph{Proc. IEEE/SICE Int. Symp. System Integration (SII)}, 2019.

\bibitem{chiu1988}
S.~L.~Chiu,
``Task compatibility of manipulator postures,''
\emph{Int. J. Robotics Research}, vol.~7, no.~5, pp.~13--21, 1988.

\bibitem{corke2017}
P.~Corke,
\emph{Robotics, Vision and Control: Fundamental Algorithms in MATLAB}, 2nd ed.
Springer, 2017.

\bibitem{coumans2021}
E.~Coumans and Y.~Bai,
``PyBullet, a Python module for physics simulation for games, robotics and machine learning,''
2016--2021. [Online]. Available: \url{http://pybullet.org}

\bibitem{ding2021}
H.~Ding, Z.~Shi, Y.~Hu, J.~Li, B.~Yu, and P.~Zhang,
``Lightweight design optimization for legs of bipedal humanoid robot,''
\emph{Structural and Multidisciplinary Optimization}, vol.~64, no.~4, pp.~2749--2762, 2021.

\bibitem{dinev2022}
T.~Dinev, C.~Mastalli, V.~Ivan, S.~Tonneau, and S.~Vijayakumar,
``A versatile co-design approach for dynamic legged robots,''
in \emph{Proc. IEEE/RSJ Int. Conf. Intelligent Robots and Systems (IROS)}, 2022.

\bibitem{fay2026}
K.~Fay, D.~A.~Djapri, A.~Zorin, J.~Clinton, A.~El~Lahib, H.~Su, M.~T.~Tolley, S.~Yi, and X.~Wang,
``House of Dextra: cross-embodied co-design for dexterous hands,''
in \emph{Proc. Int. Conf. Learning Representations (ICLR)}, 2026.

\bibitem{featherstone2008}
R.~Featherstone,
\emph{Rigid Body Dynamics Algorithms}.
Springer, 2008.

\bibitem{felis2017}
M.~L.~Felis,
``RBDL: an efficient rigid-body dynamics library using recursive algorithms,''
\emph{Autonomous Robots}, vol.~41, no.~2, pp.~495--511, 2017.

\bibitem{gosselin1991}
C.~Gosselin and J.~Angeles,
``A global performance index for the kinematic optimization of robotic manipulators,''
\emph{ASME J. Mechanical Design}, vol.~113, no.~3, pp.~220--226, 1991.

\bibitem{guri2025}
D.~Guri and G.~Kantor,
``A systematic robot design optimization methodology with application to redundant dual-arm manipulators,''
arXiv preprint arXiv:2507.21896, 2025.

\bibitem{haddadin2017}
S.~Haddadin, A.~De~Luca, and A.~Albu-Sch\"affer,
``Robot collisions: a survey on detection, isolation, and identification,''
\emph{IEEE Trans. Robotics}, vol.~33, no.~6, pp.~1292--1312, 2017.

\bibitem{iso9283}
International Organization for Standardization,
\emph{ISO 9283:1998, Manipulating industrial robots --- Performance criteria and related test methods}, 1998.

\bibitem{khatib1995}
O.~Khatib,
``Inertial properties in robotic manipulation: an object-level framework,''
\emph{Int. J. Robotics Research}, vol.~14, no.~1, pp.~19--36, 1995.

\bibitem{klein1987}
C.~A.~Klein and B.~E.~Blaho,
``Dexterity measures for the design and control of kinematically redundant manipulators,''
\emph{Int. J. Robotics Research}, vol.~6, no.~2, pp.~72--83, 1987.

\bibitem{kumar1981}
A.~Kumar and K.~J.~Waldron,
``The workspaces of a mechanical manipulator,''
\emph{ASME J. Mechanical Design}, vol.~103, no.~3, pp.~665--672, 1981.

\bibitem{li2021a}
J.~Li, J.~Liu, Y.~Hu, H.~Ding, and J.~Pang,
``Integrated optimization for service robotic arms involving workspace, drive train, structural stiffness and lightweight,''
in \emph{Proc. 5th Int. Conf. Robotics and Automation Sciences (ICRAS)}, 2021, pp.~44--50.

\bibitem{li2021b}
J.~Li, J.~Liu, H.~Ding, Y.~Hu, and J.~Pang,
``Optimization design of configuration, structure and drive train synthesis for serial robotic arms,''
in \emph{Proc. IEEE Int. Conf. Real-time Computing and Robotics (RCAR)}, 2021, pp.~945--950.

\bibitem{li2026}
J.~Li,
``Systematic lightweight method for robotics based on strain energy distribution optimization,''
arXiv preprint arXiv:2608.29146, 2026.

\bibitem{liu2024}
T.~Liu, S.~Yang, J.~Li, X.~Chen, S.~Wang, X.~Teng, W.~Lee, X.~Li, and Y.~Zheng,
``A high-performance anthropomorphic robotic arm for household applications,''
in \emph{Proc. IEEE/RSJ Int. Conf. Intelligent Robots and Systems (IROS)}, 2024, pp.~10582--10589.

\bibitem{makoviychuk2021}
V.~Makoviychuk, L.~Wawrzyniak, Y.~Guo, M.~Lu, K.~Storey, M.~Macklin, D.~Hoeller, N.~Rudin, A.~Allshire, A.~Handa, and G.~State,
``Isaac Gym: high performance GPU-based physics simulation for robot learning,''
in \emph{Proc. NeurIPS Datasets and Benchmarks Track}, 2021.

\bibitem{mingo2025}
E.~Mingo~Hoffman, D.~Costanzi, G.~Fadini, N.~Miguel, A.~Del~Prete, and L.~Marchionni,
``Addressing reachability and discrete component selection in robotic manipulator design through kineto-static bi-level optimization,''
\emph{IEEE Robotics and Automation Letters}, vol.~10, no.~3, pp.~2263--2270, 2025.

\bibitem{mittal2023}
M.~Mittal, C.~Yu, Q.~Yu, J.~Liu, N.~Rudin, D.~Hoeller, J.~L.~Yuan, R.~Singh, Y.~Guo, H.~Mazhar, A.~Mandlekar, B.~Babich, G.~State, M.~Hutter, and A.~Garg,
``Orbit: a unified simulation framework for interactive robot learning environments,''
\emph{IEEE Robotics and Automation Letters}, vol.~8, no.~6, pp.~3740--3747, 2023.

\bibitem{pahl2007}
G.~Pahl, W.~Beitz, J.~Feldhusen, and K.-H.~Grote,
\emph{Engineering Design: A Systematic Approach}, 3rd ed.
Springer, 2007.

\bibitem{paredis1993}
C.~J.~J.~Paredis and P.~K.~Khosla,
``Kinematic design of serial link manipulators from task specifications,''
\emph{Int. J. Robotics Research}, vol.~12, no.~3, pp.~274--287, 1993.

\bibitem{robodimm2026}
J.~L.~Torres, M.~Munoz, J.~D.~Alvarez, J.~L.~Blanco, and A.~Gimenez,
``Robodimm: a physics-grounded framework for automated actuator sizing in scalable modular robots,''
arXiv preprint arXiv:2603.06864, 2026.

\bibitem{salisbury1982}
J.~K.~Salisbury and J.~J.~Craig,
``Articulated hands: force control and kinematic issues,''
\emph{Int. J. Robotics Research}, vol.~1, no.~1, pp.~4--17, 1982.

\bibitem{seok2015}
S.~Seok, A.~Wang, M.~Y.~Chuah, D.~J.~Hyun, J.~Lee, D.~M.~Otten, J.~H.~Lang, and S.~Kim,
``Design principles for energy-efficient legged locomotion and implementation on the MIT Cheetah robot,''
\emph{IEEE/ASME Trans. Mechatronics}, vol.~20, no.~3, pp.~1117--1129, 2015.

\bibitem{siciliano1990}
B.~Siciliano,
``Kinematic control of redundant robot manipulators: a tutorial,''
\emph{J. Intelligent and Robotic Systems}, vol.~3, no.~3, pp.~201--212, 1990.

\bibitem{siciliano2009}
B.~Siciliano, L.~Sciavicco, L.~Villani, and G.~Oriolo,
\emph{Robotics: Modelling, Planning and Control}.
Springer, 2009.

\bibitem{strgar2025}
L.~Strgar and S.~Kriegman,
``Accelerated co-design of robots through morphological pretraining,''
arXiv preprint arXiv:2502.10862, 2025.

\bibitem{tedrake2019}
R.~Tedrake and the Drake Development Team,
``Drake: model-based design and verification for robotics,'' 2019. [Online]. Available: \url{https://drake.mit.edu}

\bibitem{thomas1985}
M.~Thomas, H.~C.~Yuan-Chou, and D.~Tesar,
``Optimal actuator sizing for robotic manipulators based on local dynamic criteria,''
\emph{ASME J. Mechanisms, Transmissions, and Automation in Design}, vol.~107, no.~2, pp.~163--169, 1985.

\bibitem{todorov2012}
E.~Todorov, T.~Erez, and Y.~Tassa,
``MuJoCo: a physics engine for model-based control,''
in \emph{Proc. IEEE/RSJ Int. Conf. Intelligent Robots and Systems (IROS)}, 2012, pp.~5026--5033.

\bibitem{vandestraete1998}
H.~J.~Van~de~Straete, P.~Degezelle, J.~De~Schutter, and R.~Belmans,
``Servo motor selection criterion for mechatronic applications,''
\emph{IEEE/ASME Trans. Mechatronics}, vol.~3, no.~1, pp.~43--50, 1998.

\bibitem{villani2018}
V.~Villani, F.~Pini, F.~Leali, and C.~Secchi,
``Survey on human--robot collaboration in industrial settings: safety, intuitive interfaces and applications,''
\emph{Mechatronics}, vol.~55, pp.~248--266, 2018.

\bibitem{wang1988}
L.-T.~Wang and B.~Ravani,
``Dynamic load carrying capacity of mechanical manipulators --- Part I: problem formulation,''
\emph{ASME J. Dynamic Systems, Measurement, and Control}, vol.~110, no.~1, pp.~46--52, 1988.

\bibitem{wensing2017}
P.~M.~Wensing, A.~Wang, S.~Seok, D.~Otten, J.~Langowski, and S.~Kim,
``Proprioceptive actuator design in the MIT Cheetah: impact mitigation and high-bandwidth physical interaction for dynamic legged robots,''
\emph{IEEE Trans. Robotics}, vol.~33, no.~3, pp.~509--522, 2017.

\bibitem{werling2021}
K.~Werling, D.~Omens, J.~Lee, I.~Exarchos, and C.~K.~Liu,
``Fast and feature-complete differentiable physics for articulated rigid bodies with contact,''
in \emph{Proc. Robotics: Science and Systems (RSS)}, 2021.

\bibitem{yoshikawa1985}
T.~Yoshikawa,
``Manipulability of robotic mechanisms,''
\emph{Int. J. Robotics Research}, vol.~4, no.~2, pp.~3--9, 1985.

\end{thebibliography}
\end{document}